\documentclass[runningheads]{llncs}
\usepackage{graphicx}
\usepackage{hyperref}
\usepackage{cite}
\usepackage{xcolor}
\usepackage{amsmath}
\usepackage{todonotes}
\usepackage{animate}
\usepackage{subcaption}
\usepackage{todonotes}

\newcommand\blfootnote[1]{%
  \begingroup
  \renewcommand\thefootnote{}\footnote{#1}%
  \addtocounter{footnote}{-1}%
  \endgroup
}

\begin{document}

\title{Temporal Consistency Objectives Regularize the Learning of Disentangled Representations}

%
\author{Gabriele Valvano\inst{1, 2} 
Agisilaos Chartsias\inst{2} \and
Andrea Leo\inst{1} \and
Sotirios A. Tsaftaris\inst{2}}


\authorrunning{G. Valvano et al.}

\titlerunning{Temporal Consistency Objectives Regularize the Learning of ...}

\institute{IMT School for Advanced Studies Lucca, Piazza S. Francesco, \\ Lucca 55100 LU, Italy
\and
School of Engineering, University of Edinburgh, West Mains Rd, \\ Edinburgh EH9 3FB, UK
}

\maketitle              
\begin{abstract}\blfootnote{Accepted for publication at: MICCAI Workshop on Domain Adaptation and Representation Transfer (DART), 2019}
There has been an increasing focus in learning interpretable feature representations,  particularly in applications such as medical image analysis that require explainability, whilst relying less on annotated data (since annotations can be tedious and costly). Here we build on recent innovations in style-content representations to learn anatomy, imaging characteristics (appearance) and temporal correlations. By introducing a self-supervised objective of predicting future cardiac phases we improve disentanglement. We propose a temporal transformer architecture that given an image conditioned on phase difference, it predicts a future frame. This forces the anatomical decomposition to be consistent with the temporal cardiac contraction in cine MRI and to have semantic meaning with less need for annotations. We demonstrate that using this regularization, we achieve competitive results and improve semi-supervised segmentation, especially when very few labelled data are available. Specifically, we show Dice increase of up to 19\% and 7\% compared to supervised
 and semi-supervised 
approaches respectively on the ACDC dataset. Code is available at: \url{https://github.com/gvalvano/sdtnet}. 

\end{abstract}

\section{Introduction}
Recent years have seen significant progress in the field of machine learning and, in particular, supervised learning. However, the success and generalization of such algorithms heavily depends learning suitable  representations \cite{bengio2013representation}. Unfortunately, obtaining them usually requires large quantities of labelled data, which need expertise and in many cases are expensive to obtain. 

It has been argued \cite{bengio2009learning} that good data representations are those separating out (disentangling) the underlying explanatory factors into disjoint subsets. As a result, latent variables become sensitive only to changes in single generating factors, while being relatively insensitive to other changes \cite{bengio2013representation}. Disentangled representations have been reported to be less sensitive to nuisance variables and to produce better generalization \cite{van2019disentangled}. In the context of medical imaging, such representations offer: i) better interpretability of the extracted features; ii) better generalization on unseen data; iii) and the potential for semi-supervised learning \cite{chartsias2019factorised}. Moreover,  disentanglement allows interpretable latent code manipulation, which is desirable in a variety of applications, such as modality transfer and multi-modal registration \cite{chartsias2019factorised, qin2019unsupervised, lee2018diverse}.

Medical images typically present the spatial information about the patient's anatomy (shapes) modulated by modality-specific characteristics (appearance). The SDNet framework \cite{chartsias2019factorised} is an attempt to decouple anatomical factors from their appearance towards more explainable representations. Building on this concept, we introduce a new architecture that drives the model to learn anatomical factors that are both spatially and temporally consistent. We propose a new model, namely: Spatial Decomposition and Transformation Network (SDTNet). 

The main \textbf{contributions} of this paper are: \textbf{(1)} we introduce a modality invariant transformer that, conditioned on the temporal information, predicts future anatomical factors from the current ones; \textbf{(2)} we show that the transformer provides a self-supervised signal useful to improve the generalization capabilities of the model;  \textbf{(3)} we achieve state of the art performance compared to SDNet for semi-supervised segmentation at several proportions of labelled data available; \textbf{(4)} and show for the first time preliminary results of cardiac temporal synthesis.

\section{Related Works}\label{sec_rel_works}

\subsection{Learning good representations with temporal conditioning} 

The world surrounding us is typically affected by smooth temporal variations and is known that temporal consistency plays a key role for the development of invariant representations in biological vision \cite{wood2016smoothness}. However, despite that temporal correlations have been used to learn/propagate segmentations in medical imaging \cite{qin2018joint, bai2018recurrent}, their use as a learning signal to improve representations remains unexplored. To the best of our knowledge, this is the first work to use spatiotemporal dynamics to improve disentangled representations in cardiac imaging.

Outside the medical imaging community, we find some commonalities of our work with Hsieh et al. \cite{hsieh2018learning}, who address the challenge of video frame prediction decomposing a video representation in a time-invariant \textit{content} vector and a time-dependent \textit{pose} vector. Assuming that the content vector is fixed for all frames, the network aims to learn the dynamics of the low-dimensional pose vector. The predicted pose vector can be decoded together with the fixed content features to generate a future video frame in pixel space. Similarly, we decompose the features space in a fixed and a time-dependent subset (\textit{modality} and \textit{anatomy}). However, our objective is not merely predicting a future temporal frame, but we use the temporal prediction as a self-supervised signal to ameliorate the quality of the representation: i.e. we constrain its temporal transformation to be smooth. By doing so, we demonstrate that we can consistently improve the segmentation capabilities of the considered baselines.

\subsection{Spatial Decomposition Network (SDNet)}\label{subsec_sdnet}

Here, we briefly review a recent approach for learning disentangled anatomy-modality representations in cardiac imaging, upon which we build our model.

The SDNet \cite{chartsias2019factorised} can be seen as an autoencoder taking as input a 2D image $x \sim X$ and decomposing it into its anatomical components $s = f_{A}(x)$ and modality components $z = f_{M}(x)$. The vector $z$ is modelled as a probability distribution $Q(z|X)$ that is encouraged to follow a multivariate Gaussian, as in the VAE framework \cite{kingma2013auto}. $s$ is a multi-channel output composed of binary discrete maps. A decoder $g(\cdot)$ uses both $s$ and $z$ to reconstruct the input image $\tilde{x} = g(s, z) \approx x $. An additional network $h(\cdot)$ is supervisedly trained to extract the heart segmentation $\tilde{y}=h(s)$ from $s$, while an adversarial signal forces $\tilde{y}$ to be realistic even when few pairs of labelled data are available, enabling semi-supervised learning. 

While SDNet was shown to achieve impressive results in semi-supervised learning, it still requires human annotations to learn to decouple the cardiac anatomy from other anatomical factors. Furthermore, it doesn't take advantage of any temporal information to learn better anatomical factors: as a result they are not guaranteed to be temporally correlated.

\section{Proposed Approach}
\label{sec_approach}
\begin{figure}[t]
    \centering
    \includegraphics[width=0.85\textwidth]{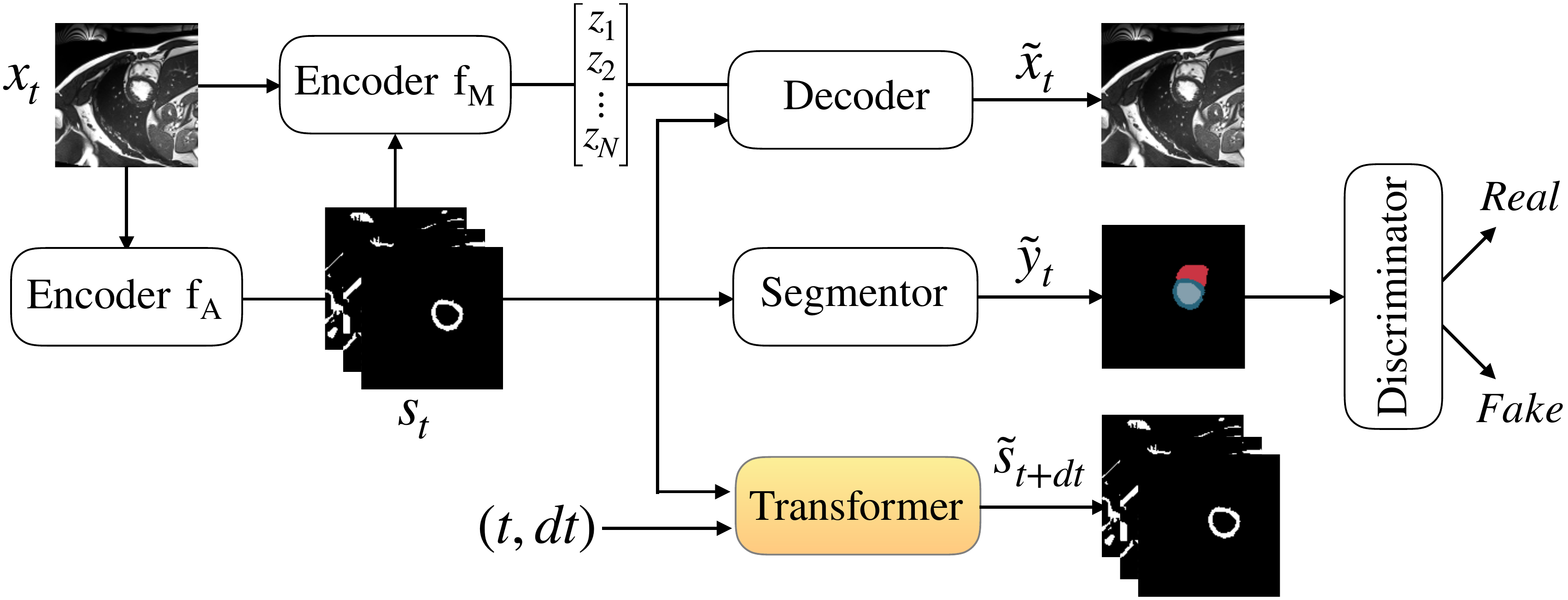}
    \caption{SDTNet block diagram. The transformer (in yellow) predicts the future anatomical factors conditioned on the temporal information. The future frame can be generated by the decoder using $\tilde{s}_{t+dt}$ and the current $z$ factor.}
    \label{fig_sdtnet}
\end{figure}

Herein, we address the above limitations, by a simple hypothesis: components $s$ of different cardiac phases should be similar within the same cardiac cycle and their differences, if any, should be consistent across different subjects.  To achieve this we introduce a new neural network $T(\cdot)$ in the SDNet framework that, conditioned on temporal information, regularizes the anatomical factors such that they can be consistent (e.g. have smooth transformations) across time. Obtaining better representations will ultimately allow improved performance in the segmentation task, too. $T(\cdot)$ is a modality-invariant transformer that `warps' the $s$ factors learnt by the SDNet according to the cardiac phase. Furthermore, by combining the current $z$ factors with the predicted $s$ factors for future time points, one can reconstruct the future frames in a cardiac sequence: e.g., given time $t_1 < t_2$, we have $\tilde{x}_{t_2} = g(T(s_{t_1}), z_{t_1}) \approx x_{t_2}$. 
Our model is shown in Figure \ref{fig_sdtnet}. Below we focus our discussion on the design of the transformer and the training costs, all other network architectures follow that of SDNet \cite{chartsias2019factorised}. In the following, $t$, $dt$ are scalars, while remaining variables are considered as tensors.

\subsection{Spatial Decomposition and Transformation Network (SDTNet)} 
\begin{figure}[t]
    \centering
    \includegraphics[width=0.90\textwidth]{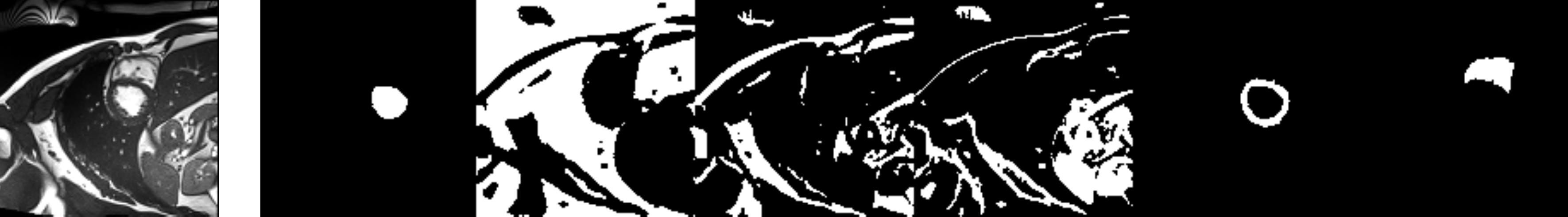}
    \caption{Anatomical factors extracted by the SDTNet from the image on the left.}
    \label{fig_decomposition}
\end{figure}

The transformer $T(\cdot)$ takes as input the binary anatomical factors $s$ (Figure \ref{fig_decomposition}) and their associated temporal information $t$. Under the assumption that the modality factors remain constant throughout the temporal dimension (e.g. the heart contracting from extra-diastole to extra-systole), the transformer must deform the current anatomy $s_{t}$ such that, given a temporal change $dt$, it estimates $s_{t + dt}$, ie. the anatomy of image $x_{t + dt}$ when given as input. Using this prediction $\tilde{s}_{t + dt}=T(s_{t}, t, dt)$ together with the fixed modality factors $z_{t}$, we should be able to correctly reconstruct the image at the future time point $\tilde{x}_{t + dt}$. By capturing the temporal dynamics of the anatomical factors, the transformer guides their generation to be temporally coherent, resulting in a self-supervised training signal, that is the prediction error of future anatomical factors.

\subsection{Transformer design}
After testing several architecture designs for the transformer, we found that the best results could be obtained by adapting a UNet \cite{ronneberger2015u} to work with binary input/output conditioned on temporal information on the bottleneck. 

Temporal information, the tuple $(t, dt)$, is encoded via an MLP consisting of 3 fully connected layers, arranged as 128-128-4096, with the output reshaped to $16\times 16\times 16$. This information is concatenated at the bottleneck of the UNet where features maps have resolution $16\times 16\times 64$, to condition the transformer and control the required deformation. To encourage the use of the temporal features and retain the notion of the binary inputs, the features at the bottleneck and of the MLP are bounded in [0, 1], using a sigmoid activation function.

We hypothesised that it would be easier to model differential changes to anatomy factors. Thus, we added a long residual connection between the UNet input and its output. We motivate this by observing that the anatomical structure that mostly changes in time is the heart: thus learning the spatial transformation should be similar to learning to segment the cardiac structure in the binary tensor $s$: a task that the UNet is known to be effective at solving. 
The output of the transformer is binarized again (key for disentanglement), as in \cite{chartsias2019factorised}.

\subsection{Cost Function and Training}
The overall cost function is the following weighted sum:
\begin{equation}
    Loss = \lambda_0\cdot L_{S} + \lambda_1\cdot L_{US} + \lambda_2\cdot L_{ADV} + \lambda_3\cdot L_{TR} \;,
\end{equation}
where $\lambda_0=10$, $\lambda_1=1$ and $\lambda_2=10$ as in \cite{chartsias2019factorised}, and $\lambda_3=1$ found experimentally.

\noindent $\boldsymbol{L_S}$ is the cost associated to the supervised task (segmentation) and can be written as $ L_{S} = L_{DICE}(y, \tilde{y}) + 0.1 \cdot L_{CE}(y, \tilde{y})$, where $y$ and $\tilde{y}$ are the ground truth and predicted segmentation masks, respectively; $L_{DICE}$ is the differentiable Dice loss evaluated on left ventricle, right  ventricle and myocardium, while $L_{CE}$ is the weighted cross-entropy on these three classes plus the background (with class weights inversely proportional to the number of pixels for the class). 

\noindent $\boldsymbol{L_{US}}$ is the cost associated to the unsupervised task and can be decomposed as $L_{US} = |\tilde{x} - x| + \lambda_{KL} \cdot D_{KL}[Q(z|X)||N(0, I)] - MI(\tilde{x}, z) $. The first term is the mean absolute error between the input and the reconstruction, while the second term is the KL divergence between $Q(z|X)$ and a Normal Gaussian (with $\lambda_{KL}$=0.1). The last term is the mutual information between the reconstruction $\tilde{x}$ and the latent code $z$ and is approximated by an additional neural network, as in the InfoGAN framework \cite{chen2016infogan}. By maximizing the mutual information between the reconstruction and $z$, we prevented posterior collapse and constrained the decoder $g(\cdot)$ to effectively use the modality factors.

\noindent $\boldsymbol{L_{ADV}}$ is the adversarial loss of a Least-Squares GAN \cite{mao2018effectiveness}, used to discriminate ground truth from predicted segmentations in the unsupervised setting.

\noindent $\boldsymbol{L_{TR}}$ is the loss associated to the self-supervised signal, computed as the differentiable Dice loss between $\tilde{s}_{t+dt}$ and $s_{t+dt}$. This error serves as a proxy for the reconstruction error of future cardiac phases $| x_{t+dt} - g(T(s_{t}), z_{t})|$. In practice, we find it much easier to train $T(\cdot)$ with a loss defined in the anatomy space rather than one on the final reconstruction: in fact, the gradients used to update the network parameters can flow into $T(\cdot)$ directly from its output layer, rather than from that of the decoder $g(\cdot)$.

The model was optimized using the Exponential Moving Average (EMA): we maintained a moving average of the parameters during training, and employed their average for testing. The learning rate was scheduled to follow a triangular wave \cite{smith2017cyclical} in the range $10^{-4}$ to $10^{-5}$ with a period of 20 epochs. Both EMA and the learning rate scheduling facilitated comparisons, allowing to detect wider and more generalizable minima (hence, reducing loss fluctuations). We used Adam \cite{kingma2014adam} with an early stopping criterion on the segmentation loss of a validation set.

\section{Experiments and Discussion}\label{sec_rexperiments}

\subsection{Data and Preprocessing}

\textbf{Data.} We used ACDC data from the 2017 Automatic Cardiac Diagnosis Challenge \cite{bernard2018deep}. These are 2-dimensional cine-MR images acquired using 1.5T and 3T MR scanners from 100 patients, for which manual segmentations for the left ventricular cavity (LV), the myocardium (MYO) and the  right  ventricle (RV) are provided in correspondence to the end-systolic (ES) and end-diastolic (ED) cardiac phases. ES and ED phase instants are also provided. We used a 3-fold cross validation and randomly divided the data to obtain 70 MRI scans for training, 15 for validation and 15 for the test sets. 
 
\noindent\textbf{Preprocessing.} After removing outliers outside 5$^{th}$ and 95$^{th}$ percentiles of the pixel values, we removed the median and normalized the images on the interquartile range, centering each volume approximately around zero. 

\noindent\textbf{Training.} Since our objective was to introduce temporal consistency in the anatomical factors rather then predicting the whole cardiac cycle, we split the cine MRI sequences in two halves: i) temporal frames in the ED-ES interval; ii) temporal frames from ES to the end of the cardiac cycle. The latter frames were reversed in their temporal order, to mimic once again the cardiac contraction: as a result, we avoided the inherent uncertainty associated to the transformations of frames in the middle of the cardiac cycle. Finally, we applied 
data augmentation at run-time, consisting of rotations, translations and scaling of each 2D slice.

\begin{figure}[t]
    \centering
    \includegraphics[width=0.9\textwidth]{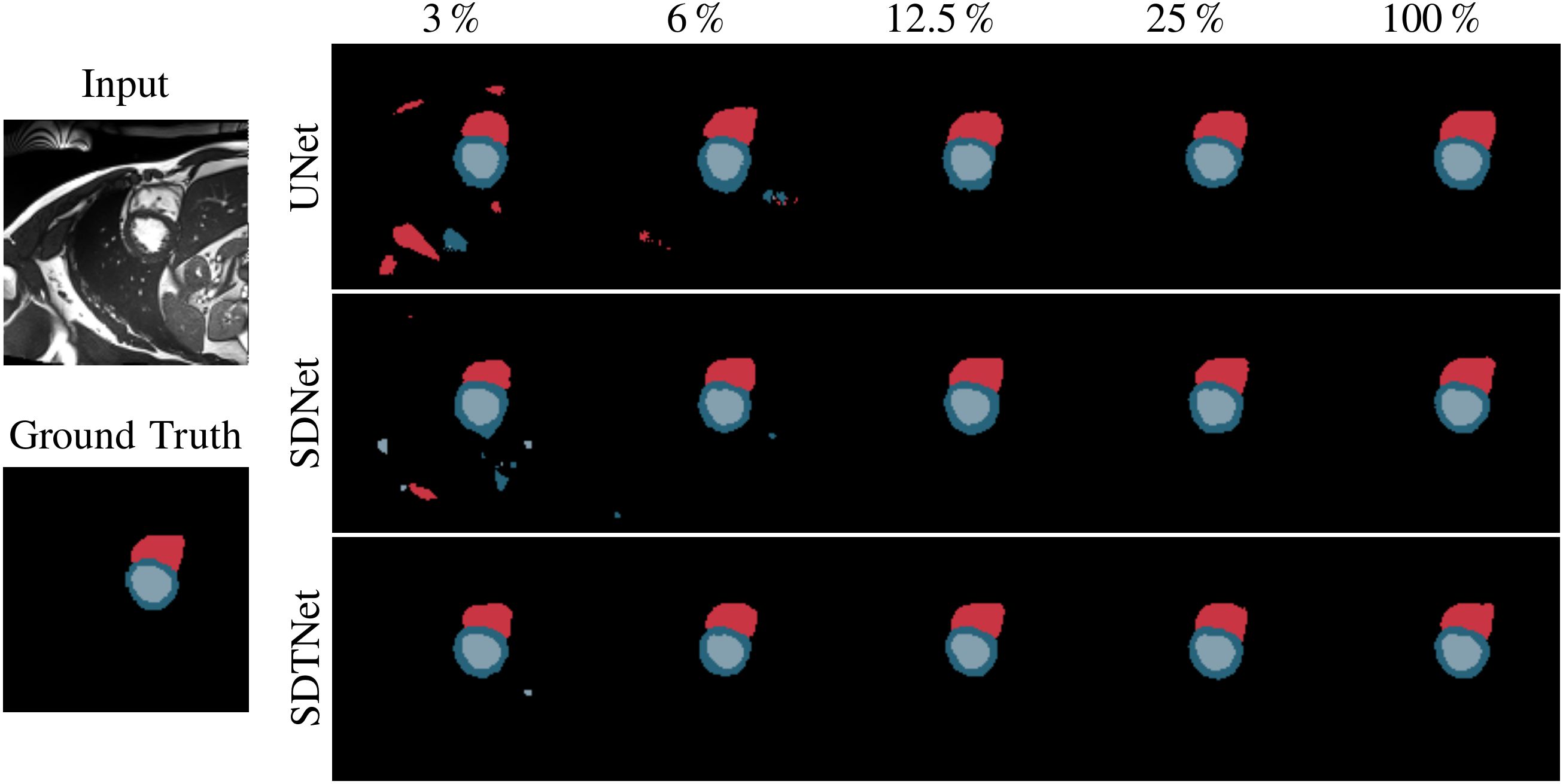}
    \caption{Comparison of predicted segmentations obtained from the UNet, SDNet, SDTNet after being trained with different percentages of the labelled data.}
    \label{fig_segmentations}
\end{figure}

\subsection{Results}

\subsubsection{Semi-supervised segmentation} We compared SDTNet to the fully supervised training of a UNet and to the semi-supervised training of SDNet in a segmentation task, varying the percentage of labelled training samples.  As Figure \ref{fig_segmentations} and Table \ref{tab_performance} show, the SDTNet consistently outperforms the others, especially at lower percentages of labelled pairs in the training set. Furthermore, SDTNet exhibits lower variance in its predictions, so it's more consistent. A paired Wilcoxon test demonstrated most of these improvements to be statistically significant. We find that the transformer forces the anatomical decomposition to follow more ``semantic'' disentanglement even with little human annotations. This translates to better segmentation results. While secondary to the thesis of the paper, both the SDNet and the SDTNet outperform the UNet.

\begin{figure}[t!]
    \centering
    \begin{subfigure}[b]{0.8\textwidth}
        \includegraphics[width=\textwidth]{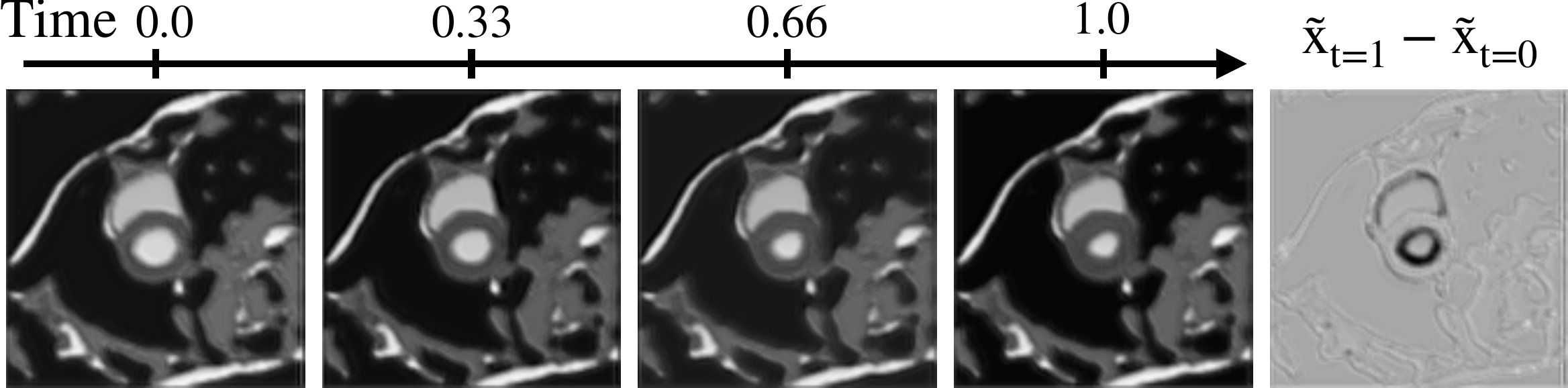}
    \end{subfigure}\hfill
    \begin{subfigure}[b]{0.15\textwidth}
        \animategraphics[width=\textwidth,loop,controls=play]{12}{foo-}{1}{7}
    \end{subfigure}
    
    \caption{
    Interpolation on the temporal axis between ED and ES phases. The images are obtained by fixing the modality-dependent factors $z_{t=0}$ and using the anatomical factors $\tilde{s}_{t>0}$ predicted for future time points. In Acrobat, clicking on the rightmost image animates frames showing the predicted cardiac contraction.
    }
    \label{fig_frame_prediction}
\end{figure}

\noindent\textbf{Cardiac synthesis} Figure \ref{fig_frame_prediction} shows that it is possible to predict future cardiac phases from ED through  ES by using the predicted anatomical factors $\tilde{s}_{t>0}$ together with the modality factors $z_{t=0}$. We note that this is the first attempt of deep learning-based temporal synthesis in cardiac albeit preliminary. Note that we train the transformer with both pathological and healthy subjects and it thus predicts average temporal transformations. Conditioning also with prior pathology information and validation of synthesis are left  as future work.

\begin{table}[t]
\setlength{\tabcolsep}{8pt} 
\renewcommand{\arraystretch}{1.0} 
\begin{center}
\begin{tabular}{l|c|c|c|c}
Labels & UNet & SDNet & SDTNet & Improvement\\
\hline
100\%   & $80.03 \ {\scriptstyle \pm 0.38} $ 
        & $85.11 \ {\scriptstyle \pm 0.73} $ 
        & $85.83 \ {\scriptstyle \pm 0.40} $ 
        & $0.72$\ \\
25\%    & $77.55 \ {\scriptstyle \pm 1.02} $
        & $81.64 \ {\scriptstyle \pm 0.96} $ 
        & $83.69 \ {\scriptstyle \pm 0.37} $ 
        & $2.05$* \\
12.5\%  & $71.04 \ {\scriptstyle \pm 1.71} $
        & $78.07 \ {\scriptstyle\pm 1.52} $ 
        & $79.48 \ {\scriptstyle \pm 0.82} $ 
        & $1.41$* \\
6\%     & $59.20 \ {\scriptstyle\pm 1.38} $  
        & $72.18 \ {\scriptstyle\pm1.91}$ 
        & $74.22 \ {\scriptstyle\pm0.57}$ 
        & $2.04$* \\
3\%     & $44.89 \ {\scriptstyle\pm 9.52} $ 
        & $56.89 \ {\scriptstyle\pm 2.48} $ 
        & $63.74 \ {\scriptstyle\pm 1.59} $ 
        & $6.85$* \\
\end{tabular}
\end{center}

\caption{DICE scores comparing SDTNet and other baselines at various proportions of available labeled data. The last column shows the average improvement of SDTNet over SDNet. Asterisks denote statistical significance ($p<0.01$).}
\label{tab_performance}
\end{table}

\section{Conclusion}\label{sec_conclusion}
We introduced a self-supervised objective for learning disentangled anatomy-modality representations in cardiac imaging. By leveraging the temporal information contained in cine MRI, we introduced a spatiotemporal model in SDNet \cite{chartsias2019factorised}, improving its generalization capabilities in the semi-supervised setting at several proportions of labelled data available. Also, the resulting approach considerably outperforms the fully-supervised baseline, confirming the potential for semi-supervised and self-supervised training in medical imaging.

\subsubsection*{Acknowledgements} This work was supported by the Erasmus+ programme of the European Union, during an exchange between IMT School for Advanced Studies Lucca and the School of Engineering, University of Edinburgh. S.A. Tsaftaris acknowledges the support of the Royal Academy of Engineering and the Research Chairs and Senior Research Fellowships scheme. We thank NVIDIA Corporation for donating the Titan Xp GPU used for this research.


\bibliographystyle{splncs04}
\bibliography{Paper11} 
\end{document}